\documentclass[conference,a4paper]{IEEEtran}

\usepackage{graphicx}
\usepackage{amsmath}
\usepackage{amssymb}
\usepackage{algorithmic}
\usepackage{algorithm}
\usepackage{subfigure}
\usepackage{array}
\usepackage{tabularx}
\usepackage{multirow}

\newcommand{\bfx}{{\textbf{x}}}
\newcommand{\bfz}{{\textbf{z}}}

\begin{document}

\title{Optimizing Top Precision Performance Measure of Content-Based Image Retrieval by Learning Similarity Function}

\author{
\IEEEauthorblockN{Ru-Ze Liang}
\IEEEauthorblockA{
King Abdullah University of\\
Science and Technology,\\
Saudi Arabia\\
ruzeliang@outlook.com}
\and
\IEEEauthorblockN{Lihui Shi}
\IEEEauthorblockA{Centerfield Corporation,\\
Los Angeles, \\
CA 90245, USA\\
shilihui@uw.edu}
\and
\IEEEauthorblockN{Haoxiang Wang}
\IEEEauthorblockA{
Department of Electrical \\
and Computer Engineering,\\
Cornell University,\\
Ithaca, NY 14850, USA\\
wanghaoxiang1102@outlook.com}
\and
\IEEEauthorblockN{Jiandong Meng}
\IEEEauthorblockA{Shandong Medical College,\\
Linyi, Shangdong 276000, China\\
jiandongmeng26@hotmail.com}
\and
\IEEEauthorblockN{Jim Jing-Yan Wang}
\IEEEauthorblockA{
New York University Abu Dhabi,\\
United Arab Emirates\\
Provincial Key Laboratory for\\
Computer Information Processing\\
Technology, Soochow University,\\
Suzhou 215006, China\\
jimjywang@outlook.com
}
\and
\IEEEauthorblockN{Qingquan Sun}
\IEEEauthorblockA{
School of Computer Science\\
and Computer Engineering,\\
California State University\\
San Bernardino,\\
San Bernardino, CA 92407, USA\\
qsun@csusb.edu}
\and
\IEEEauthorblockN{Yi Gu}
\IEEEauthorblockA{
Travelers Canada,\\
Toronto, ON, Toronto\\
ygu023@gmail.com}
}

\maketitle

\begin{abstract}
In this paper we study the problem of content-based image retrieval. In this problem, the most popular performance measure is the top precision measure, and the most important component of a retrieval system is the similarity function used to compare a query image against a database image. However, up to now, there is no existing similarity learning method proposed to optimize the top precision measure. To fill this gap, in this paper, we propose a novel similarity learning method to maximize the top precision measure. We model this problem as a minimization problem with an objective function as the combination of the losses of the relevant images ranked behind the top-ranked irrelevant image, and the squared Frobenius norm of the similarity function parameter. This minimization problem is solved as a quadratic programming problem. The experiments over two benchmark data sets show the advantages of the proposed method over other similarity learning methods when the top precision is used as the performance measure.
\end{abstract}

\IEEEpeerreviewmaketitle

\section{Introduction}

In the problem of content-based image retrieval, the two basic steps of the content-based image retrieval are image visual feature extraction, and the similarity calculation. Recently, the similarity learning problem has been a hot topic in the field of machine learning. This problem tries to learn a parameterized similarity function from a training set, so that a targeted performance can be optimized \cite{Kang20151251}. To evaluate the performance of a retrieval system, common measures for retrieval performance are taken, including the area under the receiver operating characteristic (ROC) curve, and the top precision value \cite{Hong201635,Li20141502,fan2010enhanced}. The top precision measure is defined as the portion of the top-ranked relevant images out of all the relevant database images. The top-ranked relevant images are further defined as the database images which are ranked before the first ranked irrelevant database image. Top precision is an important performance measure for retrieval problems, however, all the existing similarity learning methods ignore the performance measures, and the resulted similarity function cannot give an optimal performance measure of top precision.

To fill the gap between similarity learning and top precision optimization in the problem of content-based image retrieval, we propose a problem of top precision-optimal similarity learning, together with a corresponding solution using a novel algorithm. Our contributions in this paper are divided into two parts, which are given as followed. The proposed problem is to learn a similarity function to compare a query image against a database image. With the similarity scores between query and database images, when retrieval task is performed, we can obtain a maximum top precision measure. The problem is to adjust the parameters of the similarity function, so that the top precision can be maximized. The proposed novel similarity function learning algorithm is to maximize the top precision over a training set of query images and a database of images. We design the similarity function as a linear function of the two feature vectors of a query image and a database image. To learn the parameter matrix of the similarity function, we propose to maximize the top precision for each query, and parallelly minimize the squared Frobenius norm of the parameter matrix in order to prevent the emergence of the over-fitting problem. The optimization problem is modeled by a hinge loss and solved as a quadratic programming.

The rest parts of this paper are organized as followed. In section \ref{sec:method}, we introduce the proposed similarity learning algorithm. In section \ref{sec:experiments}, we give the experimental results of the proposed method. In section \ref{sec:conclusion}, we give the conclusion of this paper with future works.

\section{Proposed similarity learning method}
\label{sec:method}

We assume we have an image database of $m$ images, $\mathcal{D}=\{\bfx_1,\cdots, \bfx_m\}$, where $\bfx_j\in \mathbb{R}^d$ is the feature vector of the $j$-th database image, a set of query images $Q=\{\bfz_1,\cdots, \bfz_n\}$, where $\bfz_i\in \mathbb{R}^d$ is the feature vector of the $i$-th query image, and a relevance matrix $Y=[y_{ij}] \in \{1,0\}^{n\times m}$, where $y_{ij} = 1$ if $\bfz_i$ is relevant to $\bfx_j$, and $0$ otherwise. We define a similarity function for a query $\bfz$ and a database image $\bfx$ parameterize by a matrix $W\in \mathbb{R}^{d\times d}$,

\begin{equation}
\label{equ:similarity}
\begin{aligned}
s(\bfz,\bfx) = \bfz^\top W \bfx.
\end{aligned}
\end{equation}

The first ranked irrelevant database image of a query $\bfz_i$, denoted as \emph{top irrelevant}, is defined as the irrelevant database image with the largest similarity to the query,

\begin{equation}
\begin{aligned}
\bfx_{\phi_i}, ~where ~\phi_i = \underset{k:y_{ik} = 0}{\arg\max}~ s(\bfz_i,\bfx_k).
\end{aligned}
\end{equation}
Top precision measure of a query $\bfz_i$ is defined as the portion of the relevant images ranked before the top irrelevant image,

\begin{equation}
\begin{aligned}
&Top~Precision (\bfz_i) \\
&= \frac{|\{\bfx_j: y_{ij} = 1,~ and~s(\bfz_i,\bfx_j) > s(\bfz_i,\bfx_{\phi_i})\}|}
{|\{\bfx_j: y_{ij} = 1\}|}.
\end{aligned}
\end{equation}
To obtain a good performance measure of top precision, we hope that for any pair of query and relevant database image, $(\bfz_i,\bfx_j)_{i,j:y_{ij}=1}$, their similarity $(\bfz_i,\bfx_j)$ is larger than the similarity to the top irrelevant database image plus a margin, $s(\bfz_i,\bfx_{\phi_i})+1$,

\begin{equation}
\label{equ:condition2}
\begin{aligned}
&s(\bfz_i,\bfx_j) > \underset{k:y_{ik} = 0}{\max} s(\bfz_i,\bfx_k) + 1,\\
&\Rightarrow 0 > \underset{k:y_{ik} = 0}{\max} s(\bfz_i,\bfx_k) - s(\bfz_i,\bfx_j) + 1, \forall~i,j:y_{ij} = 1.
\end{aligned}
\end{equation}
To maximize the top precision, we propose to learn the similarity function parameter matrix $W$ by penalizing the case which doesn't satisfies this condition,

\begin{equation}
\label{equ:condition3}
\begin{aligned}
0 \leq \underset{k:y_{ik} = 0}{\max} s(\bfz_i,\bfx_k) - s(\bfz_i,\bfx_j) + 1. \forall~i,j:y_{ij} = 1.
\end{aligned}
\end{equation}
Thus the loss function for each $(\bfz_i,\bfx_j)_{i,j:y_{ij} = 1}$ is,

\begin{equation}
\label{equ:condition}
\begin{aligned}
\ell(\bfz_i,\bfx_j; W)&=\max\left(0, \underset{k:y_{ik} = 0}{\max} s(\bfz_i,\bfx_k) - s(\bfz_i,\bfx_j) +1\right )\\
&=\max\left(0, \underset{k:y_{ik} = 0}{\max} \left ( \bfz_i^\top W (\bfx_k - \bfx_j) +1 \right )\right ).
\end{aligned}
\end{equation}
The optimization problem is modeled as,

\begin{equation}
\label{equ:objective2}
\begin{aligned}
\min_{W,\xi_{ij}|_{i,j:y_{ij}=1}} ~&\left \{ \frac{1}{2}\|W\|_F^2+C\sum_{i,j:y_{ij}=1} \xi_{ij} \right \},\\
s.t.~&\forall i,j:y_{ij}=1,\\
&\xi_{ij}\geq 0, \\
&\xi_{ij}\geq \bfz_i^\top W (\bfx_k - \bfx_j) +1,
\end{aligned}
\end{equation}
where $\xi_{ij}$ is slack variable of (\ref{equ:condition}), $\frac{1}{2}\|W\|_F^2$ is the squared Frobenius norm of $W$ to prevent the over-fitting problem, and $C$ is the tradeoff parameter.

%
%

The Lagrange function of (\ref{equ:objective2}) is
\begin{equation}
\begin{aligned}
\mathcal{L}=
&\frac{1}{2}\|W\|_F^2+C\sum_{i,j:y_{ij}=1} \xi_{ij}\\
&-\sum_{i,j:y_{ij}=1}\alpha_{ij}\xi_{ij}\\
&-\sum_{i,j,k:y_{ij}=1,y_{ik}=0} \beta_{ijk}\left (
\xi_{ij}- \bfz_i^\top W (\bfx_k - \bfx_j) -1\right ),
\end{aligned}
\end{equation}
where $\alpha_{ij}\geq 0$ is the Lagrange multiplier of the constraint $\xi_{ij}\geq 0$, and $\beta_{ijk}\geq 0$ is the Lagrange multiplier of the constraint $\xi_{ij}\geq \bfz_i^\top W (\bfx_k - \bfx_j) +1$. The dual form of the optimization problem of (\ref{equ:objective2}) is given as follows,

\begin{equation}
\label{equ:lagrange}
\begin{aligned}
&\max_{\alpha_{ij}|_{i,j:y_{ij}=1},\beta_{ijk}|_{i,j,k:y_{ij}=1,y_{ik}=0}}
\min_{W,\xi_{ij}|_{i,j:y_{ij}=1}}\mathcal{L}\\
&s.t.\alpha_{ij}\geq0, \forall i,j:y_{ij}=1,\\
&~~~\beta_{ijk}\geq 0, \forall i,j,k:y_{ij}=1,y_{ik}=0.
\end{aligned}
\end{equation}
By setting the derivative of $\mathcal{L}$ with regard to $\xi_{ij}$ to zero, we have

\begin{equation}
\label{equ:xi}
\begin{aligned}
&\frac{\partial \mathcal{L}}{\partial \xi_{ij}}=
C-\alpha_{ij}-\sum_{k:y_{ik}=0} \beta_{ijk}=0,\\
&\Rightarrow \alpha_{ij}= C-\sum_{k:y_{ik}=0} \beta_{ijk}\geq 0,\\
&\Rightarrow  C\geq \sum_{k:y_{ik}=0} \beta_{ijk}.
\end{aligned}
\end{equation}
By substituting (\ref{equ:xi}) to $\mathcal{L}$, we have

\begin{equation}
\label{equ:L1}
\begin{aligned}
\mathcal{L}
=&\frac{1}{2}\|W\|_F^2-\sum_{i,j,k:y_{ij}=1,y_{ik}=0} \beta_{ijk}
\bfz_i^\top W (\bfx_j - \bfx_k)\\
&+ \sum_{i,j,k:y_{ij}=1,y_{ik}=0} \beta_{ijk}.
\end{aligned}
\end{equation}
By setting the derivative of $\mathcal{L}$ with regard to $W$ to zero, we have

\begin{equation}
\label{equ:W}
\begin{aligned}
&\frac{\partial \mathcal{L}}{\partial W}=
W-\sum_{i,j,k:y_{ij}=1,y_{ik}=0} \beta_{ijk} \bfz_i (\bfx_j - \bfx_k)^\top =0\\
&\Rightarrow
W=\sum_{i,j,k:y_{ij}=1,y_{ik}=0} \beta_{ijk} \bfz_i (\bfx_j - \bfx_k)^\top.
\end{aligned}
\end{equation}
By substituting $W$ of (\ref{equ:W}) to $\mathcal{L}$ of (\ref{equ:L1}), we have

\begin{equation}
\begin{aligned}
\mathcal{L}
=&-\frac{1}{2}\underset{i,j,k: y_{ij}=1,y_{ik}=0}{\sum} \sum_{i',j',k':y_{i'j'}=1,y_{i'k'}=0}
\beta_{ijk} \beta_{i'j'k'}
\bfz_i^\top \bfz_{i'}  \\
&(\bfx_{j'} - \bfx_{k'})^\top(\bfx_j- \bfx_k) + \sum_{i,j,k:y_{ij}=1,y_{ik}=0} \beta_{ijk}.
\end{aligned}
\end{equation}
This is a quadratic function of the multiplier variables, $\beta_{ijk}|_{i,j,k:y_{ij}=1,y_{ik}=0}$. The optimization problem of (\ref{equ:lagrange}) is transferred to

\begin{equation}
\label{equ:QP}
\begin{aligned}
\max_{\beta_{ijk}|_{i,j,k:y_{ij}=1,y_{ik}=0}}
~&\left \{ -\frac{1}{2}\underset{i,j,k: y_{ij}=1,y_{ik}=0}{\sum} \sum_{i',j',k':y_{i'j'}=1,y_{i'k'}=0}
 \right.\\
& \beta_{ijk} \beta_{i'j'k'}
\bfz_i^\top \bfz_{i'} (\bfx_{j'} - \bfx_{k'})^\top(\bfx_j - \bfx_k)\\
&\left . + \sum_{i,j,k:y_{ij}=1,y_{ik}=0} \beta_{ijk}\right \},\\
s.t.~&
C\geq \sum_{k:y_{ik}=0} \beta_{ijk}, ~and~\beta_{ijk}\geq 0,\\
& \forall ~i,j,k:y_{ij}=1,y_{ik}=0.
\end{aligned}
\end{equation}
This is a quadratic programming (QP) problem, and we can use a standard active set algorithm to solve this problem. After the $\beta_{ijk}|_{i,j,k:y_{ij}=1,y_{ik}=0}$ are solved, we can recover $W$ from (\ref{equ:W}).

\section{Experiments}
\label{sec:experiments}

\subsection{Experimental protocol}

In this experiment, we used two benchmark image databases, which are Indoor image database \cite{Quattoni2009413}, and Caltech256 image database \cite{griffin2007caltech}. Then total number of images/  number classes of these two databases are 15,620/ 67 and 30,670/ 256.
Given a database, we first split it to a set of queries and a set of database images, and these queries are further split into  a training set and a test set. We apply the proposed similarity learning algorithm to  the training set of queries and the database images to learn the similarity function. Then we use the similarity function to conduct the retrieval task for the test set of queries. The performance is evaluated by top precision measure.

\subsection{Experimental results}

\begin{figure}
\centering
\includegraphics[width=0.4\textwidth]{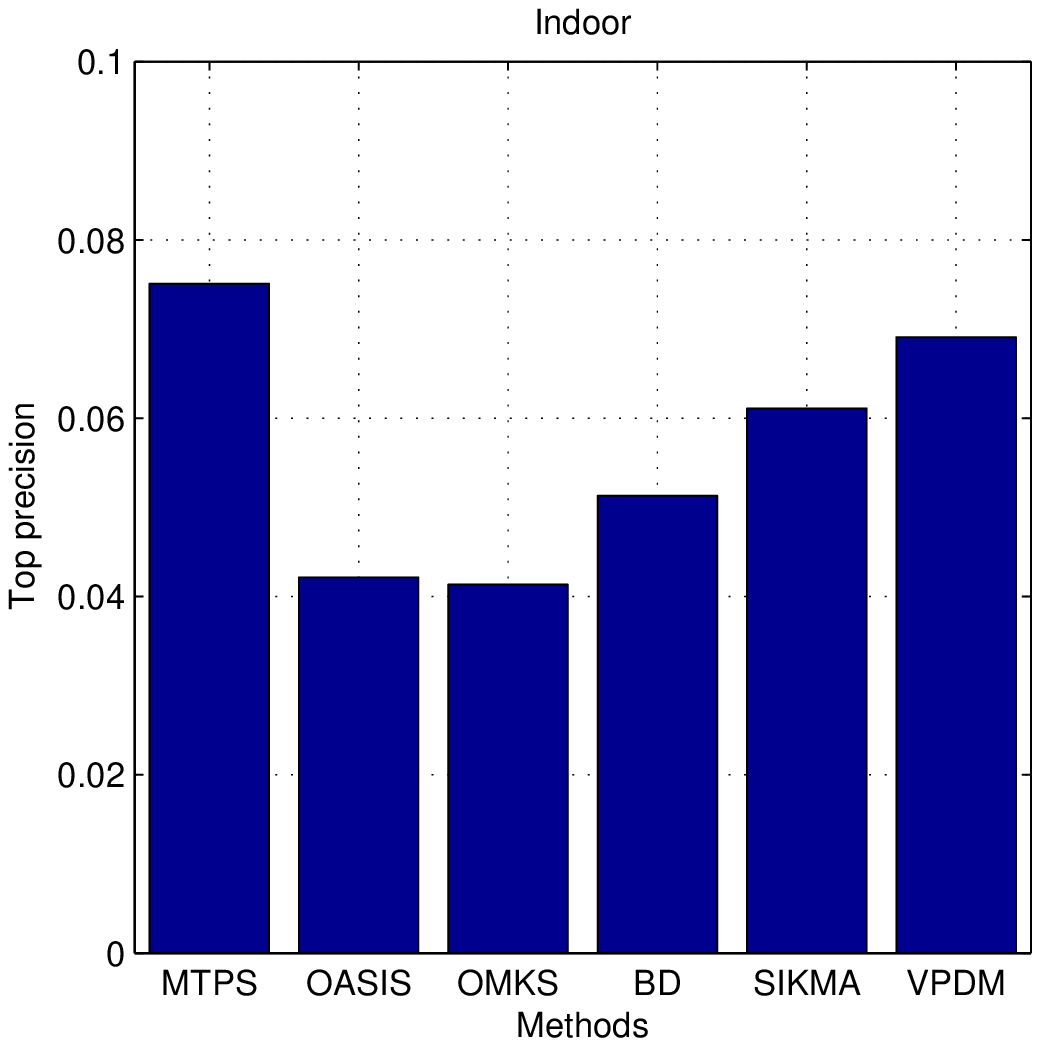}
\includegraphics[width=0.4\textwidth]{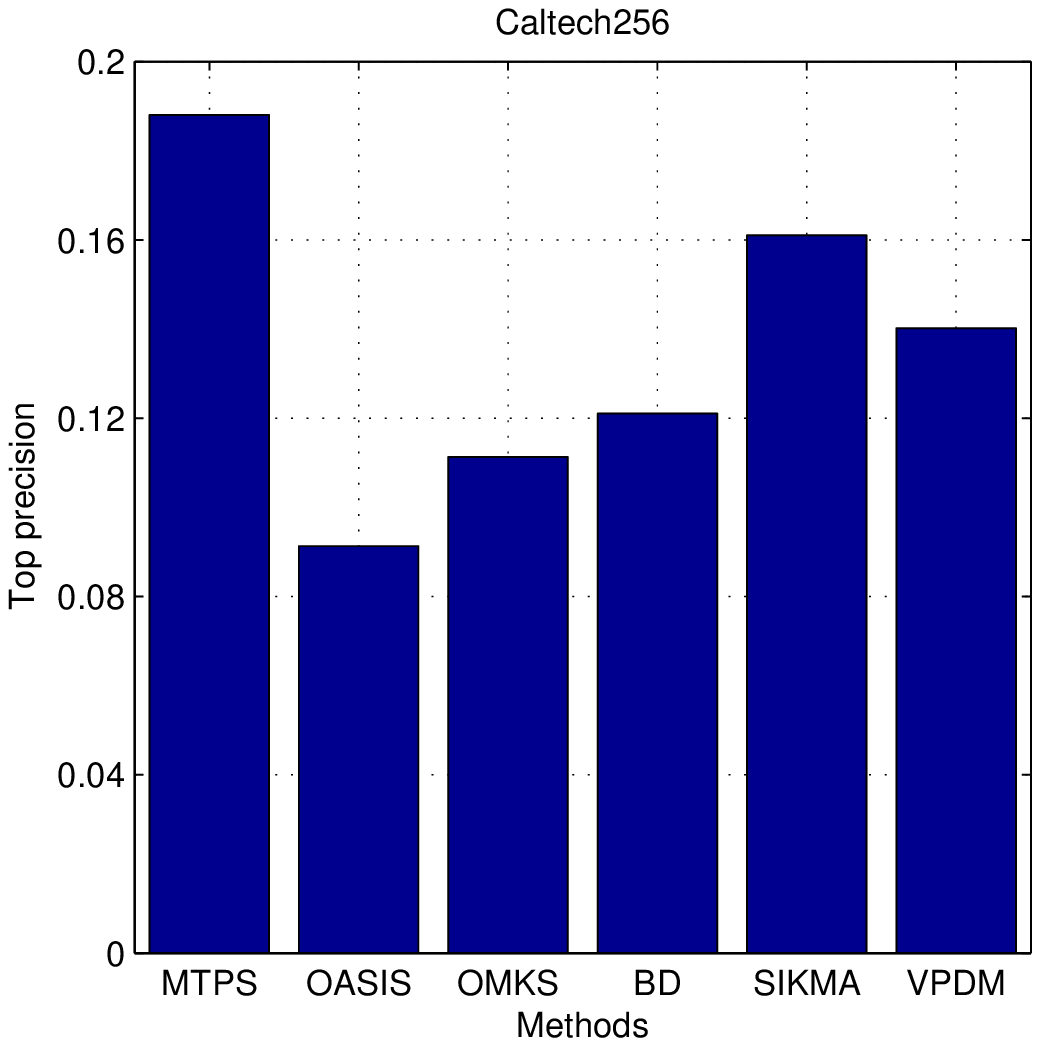}
\caption{Comparison of different similarity learning algorithms for retrieval task over two benchmark databases.}
\label{Fig:result}
\end{figure}

We compare our proposed maximum top precision similarity learning method (MTPS), to several state-of-the-art similarity learning methods,
 online algorithm for scalable image similarity (OASIS) \cite{Chechik200911},
 online multiple kernel similarity (OMKS) \cite{Xia2014536},
 Boosted distance (BD) \cite{Yu2008451},
 stochastic intersection kernel machine (SIKMA) \cite{Wang20122177}, and
 visuality-preserving distance metric (VPDM) \cite{Yang201030}.
The results of the comparison over the four benchmark databases are given in Figure \ref{Fig:result}. It is obvious that the proposed algorithm, MTPS, outperforms the compared similarity learning algorithms over both the two benchmark databases. In the results over the Caltech256, the top precisions of almost all the compared algorithms are below 0.16, while our algorithms, MTPS, have top precisions higher than 0.18.

\begin{figure}
\centering
\includegraphics[width=0.4\textwidth]{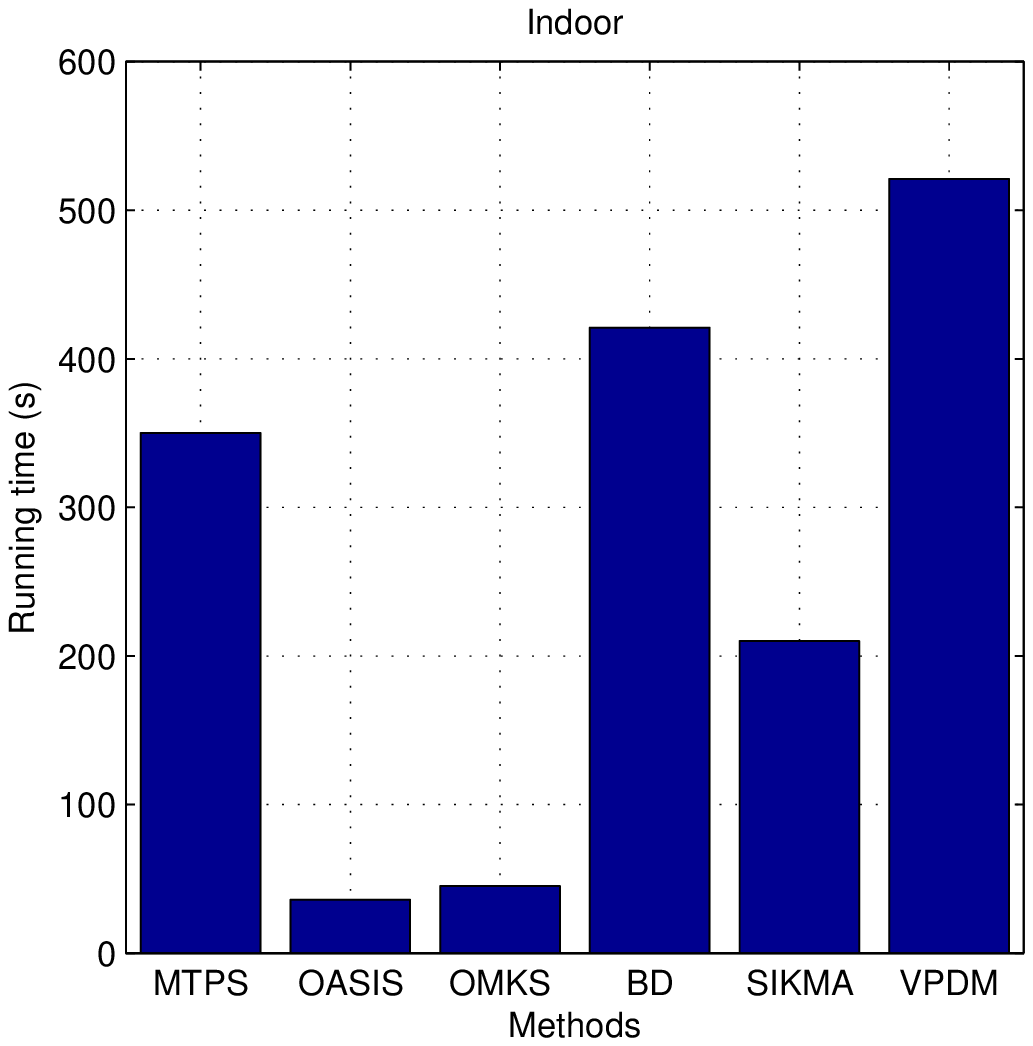}
\includegraphics[width=0.4\textwidth]{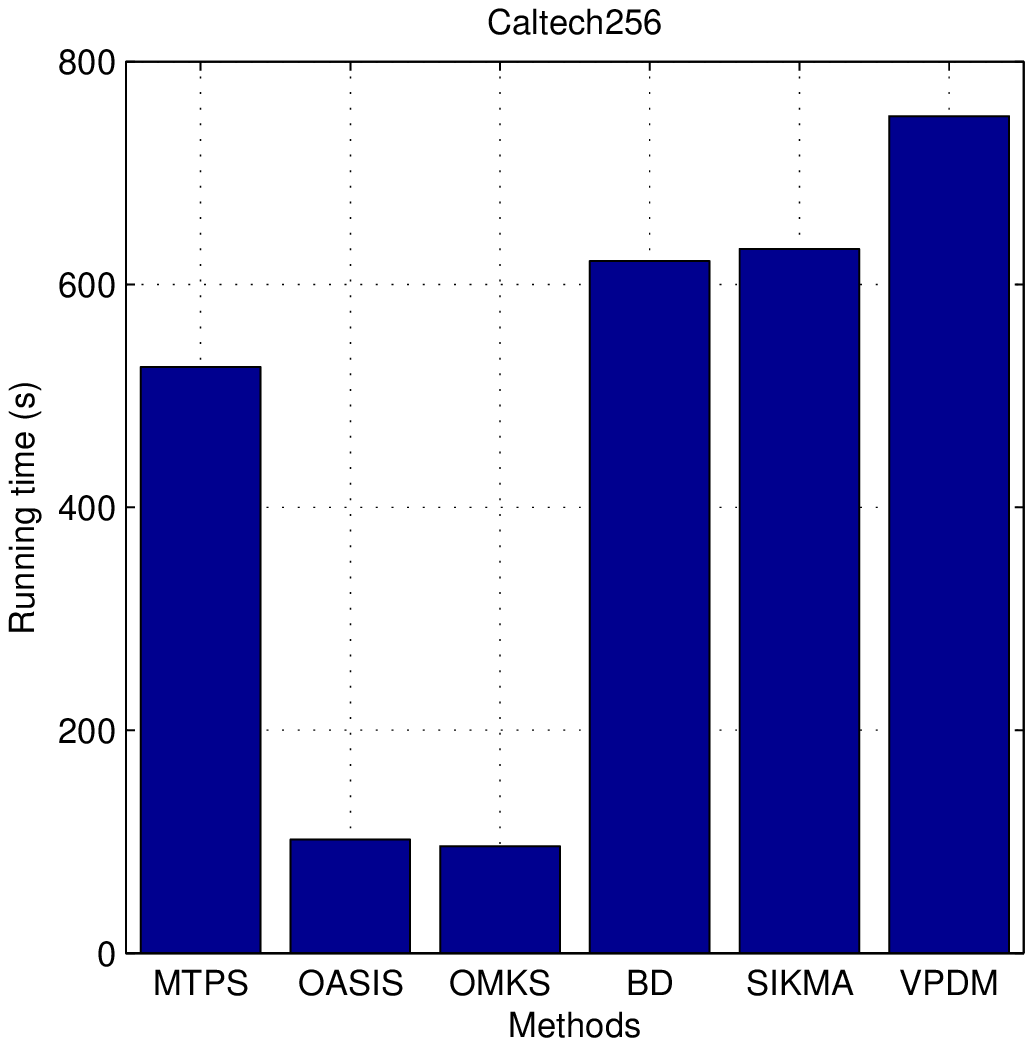}
\caption{Running of different similarity learning algorithms over two benchmark databases.}
\label{Fig:time}
\end{figure}

In addition, we test the running time of the proposed algorithm over the two benchmark databases, displayed in Figure \ref{Fig:time}. Our proposed algorithm MTPS exhibits shorter running time than others, except for the OASIS and OMKS, suggesting that MTPS provides a comparably efficient way against the problem, and shows a promising improvement space for the future work.


\section{Conclusion and future works}
\label{sec:conclusion}

In this paper we study the problem of similarity learning for content-based image retrieval. The critical novelty of this work is it learns the similarity to maximize the top precision measure over the training set. We model the learning problem as a support top irrelevant database image weighting problem to obtain the optimal similarity function. The experiments over some benchmark databases show its advantages over other similarity learning methods. In the future, we will investigate to use some other similarity function as similarity measure instead of linear function, such as Bayesian network \cite{fan2014tightening,fan2014finding,fan2015improved}, and also to develop novel algorithms of other machine learning problems and applications besides similarity learning, to maximize top precision measure, such as importance sampling \cite{shi2009new,shi2015synthesis,fan2011margin}, portfolio choices \cite{shen2015portfolio,shen2016portfolio}, multimedia technology \cite{ma2014design,xu2015coupled,lin2016multi,liu2015supervised,wang2015multiple,wang2015supervised,wang2014effective,wang2015representing,wang2015image,xu2016mechanical}, computational biology \cite{wang2014computational,zhou2014biomarker,liu2013structure,peng2015modeling,wang2015remote}, big data processing \cite{wang2015towards,wang2014next,zhao2014fusionfs,wang2014optimizing,li2013zht}, computer vision \cite{guo2012study,wang2014dog,wang2014leveraging,wang2015leveraging,dingmanifoldlearning,generalizedGaussian,wang2015computational,wang2015unsupervised,wang2015evaluation,yang2015visual,shen2015learning,king2015surgical,li2015outlier,li2015burn}, information security \cite{wang2007gradual,luo2014federated,xu2014evasion,xu2013cross,xu2014adaptive,xu2012push}, and medical imaging \cite{hu2014inferring,hu2015spectral,hu2016matched},
etc.

\section*{Acknowledgements}

The study is supported by a grant from Provincial Key Laboratory
for Computer Information Processing Technology, Soochow University, China.



\end{document}